\documentclass[lettersize,journal,compsoc]{IEEEtran}
\usepackage{amsmath,amsfonts,amssymb}
\usepackage{array}
\usepackage[caption=false,font=normalsize,labelfont=sf,textfont=sf]{subfig}
\usepackage{textcomp}
\usepackage{stfloats}
\usepackage{url}
\usepackage{verbatim}
\usepackage{graphicx}
\usepackage{cite}
\usepackage{booktabs}

\usepackage{epsfig}
\usepackage{subfig}
\usepackage{caption}
\usepackage{longtable}
\usepackage{multirow}
\usepackage{algorithmicx}
\usepackage{algpseudocode}
\usepackage{indentfirst} 
\usepackage{color} 
\usepackage{threeparttable}
\usepackage{tablefootnote}
\usepackage{diagbox}
\usepackage{booktabs}
\usepackage[switch]{lineno}

\definecolor{LightCyan}{rgb}{0.88,1,1}
\definecolor{LightGray}{rgb}{0.82, 0.82, 0.82}

\usepackage{xcolor,colortbl}
\usepackage{pifont}
\usepackage{bm}
\usepackage[ruled]{algorithm2e}
\usepackage{mathrsfs}
\usepackage{amsmath}

\begin{document}

\title{Gradient Projection For Continual Parameter- \\ Efficient Tuning}

\author{Jingyang Qiao,
Zhizhong Zhang,
Xin Tan,
Yanyun Qu,
Wensheng Zhang,
Zhi Han, \\
Yuan Xie,~\IEEEmembership{~Member,~IEEE}
\IEEEcompsocitemizethanks{
\IEEEcompsocthanksitem J. Qiao, Z. Zhang, X. Tan and Y. Xie are with School of Computer Science and Technology, East China Normal University, Shanghai, 200062, China; E-mail: 52275901010@stu.ecnu.edu.cn, \{zzzhang, xtan, yxie\}@cs.ecnu.edu.cn
\IEEEcompsocthanksitem Y. Qu is with School of Information Science and Technology, Xiamen University, Fujian, 361005, China; E-mail: yyqu@xmu.edu.cn
\IEEEcompsocthanksitem W. Zhang is with Institute of Automation, Chinese Academy of Sciences, Beijing, 100190, China; E-mail: wensheng.zhang@ia.ac.cn
\IEEEcompsocthanksitem Z. Han is with Shenyang Institute of Automation, Chinese Academy of Sciences, Shenyang, 110016, China; E-mail: hanzhi@sia.cn}


\thanks{}
\thanks{}}

\IEEEtitleabstractindextext{

\begin{abstract}
Parameter-efficient tunings (PETs) have demonstrated impressive performance and promising perspectives in training large models, while they are still confronted with a common problem: the trade-off between learning new content and protecting old knowledge, leading to zero-shot generalization collapse, and cross-modal hallucination. In this paper, we reformulate Adapter, LoRA, Prefix-tuning, and Prompt-tuning from the perspective of gradient projection, and firstly propose a unified framework called \underline{\textbf{P}}arameter \underline{\textbf{E}}fficient \underline{\textbf{G}}radient \underline{\textbf{P}}rojection (PEGP). We introduce orthogonal gradient projection into different PET paradigms and theoretically demonstrate that the orthogonal condition for the gradient can effectively resist forgetting even for large-scale models. It therefore modifies the gradient towards the direction that has less impact on the old feature space, with less extra memory space and training time. We extensively evaluate our method with different backbones, including ViT and CLIP, on diverse datasets, and experiments comprehensively demonstrate its efficiency in reducing forgetting in class, online class, domain, task, and multi-modality continual settings. The project page is available at https://dmcv-ecnu-pegp.github.io/.
\end{abstract}

}

\maketitle

\section{Introduction}
\IEEEPARstart{L}{arge} pre-trained models currently are the prevailing research area in artificial intelligence. In this framework, parameter-efficient tunings (PETs) provide a new perspective for downstream adaption and enable the large pre-trained model to be implemented in specialized domains \cite{Prompt, Prefix, Adapter, LoRA}. However, when continually adapting to consecutive downstream tasks, PETs still exhibit catastrophic forgetting, the phenomenon of learning the current knowledge while concurrently eroding the preceding memory \cite{CF1, CF2, CF3, Learn}, \textit{e.g.}, losing the zero-shot generalization ability or producing hallucination.

In seeking to address the forgetting problem, continual learning, or incremental learning \cite{CLS1, CLS2, CLS3, CLS4} is proposed to train a model with continuously expanded datasets by adding novel classes or domains \cite{ICaRL, DER++, EWC, OWM, AGEM, DER, Foster, Packnet, HAT, Selective, Dynamically}. However, traditional continual learning models are unsuitable for fine-tuning large models, \textit{i.e.}, we can’t replay or expand as data/model is huge. Additionally, due to the limited connection between distinct PETs, there lacks a unified solution for PETs \cite{L2P, DualPrompt, CODA, APrompts}. 

Fortunately, gradient projection (GP), regardless of the backbone network, indicates that continual learning would not forget if the gradient is updated in the orthogonal direction to the subspace spanned by the old features \cite{GPM, NSCL, OGD}. It modifies the gradient into an orthogonal direction by gradient projection matrix, extracting from sampled feature space. It can provide the anti-forgetting mechanism consuming with fewer extra memory space and training time, which can reduce the burden of fine-tuning large models. However, GP is based on convolutional neural networks \cite{Resnet}, leading to the present GP theory cannot be directly applied to parameter-efficient tuning methods. Notice that our prior work has demonstrated that GP theory can be combined with Prompt/Prefix-tuning paradigms \cite{PGP}, while it still lacks a unified anti-forgetting framework with the PET-based methods.

Motivated by the preceding analysis, in this paper, we propose a novel \textbf{P}arameter \textbf{E}fficient \textbf{G}radient \textbf{P}rojection (PEGP) method. We recall the pipeline of distinct parameter-efficient tuning (\textit{i.e.}, Prompt-tuning, Prefix-tuning, Adapter, and LoRA) and find that they own a union anti-forgetting equation. Moreover, we solve this anti-forgetting equation by conducting Singular Value Decomposition (SVD) on a sampled feature space. That allows us to obtain the gradient projection matrix in an efficient way. 

\begin{figure*}[htb]
\centering
\includegraphics[width=7in]{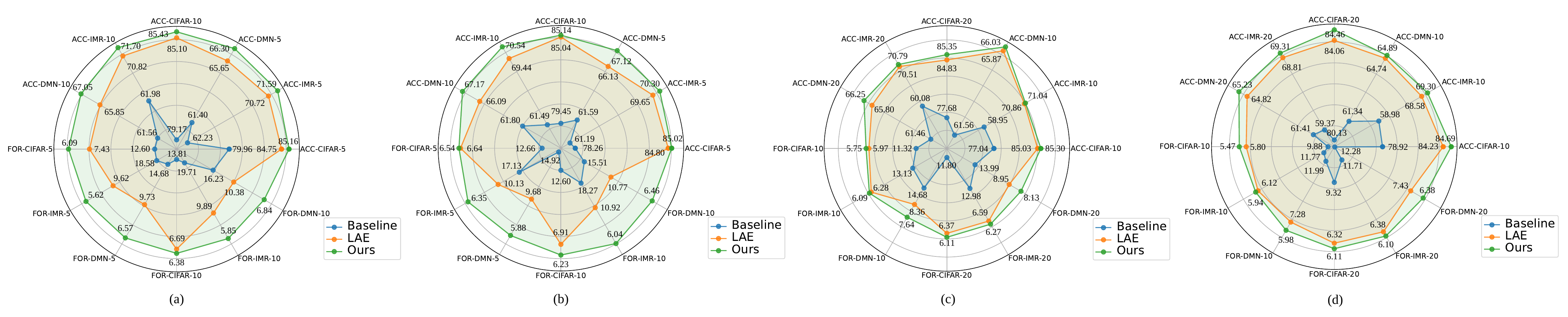}
\caption{Radar chart of continual learning results on multiple datasets based on ViT backbone with (a) Adapter (b) LoRA (c) Prefix-Tuning (d) Prompt-Tuning paradigms. ACC refers to the average accuracy metric (higher is better). FOR refers to the forgetting metric (lower is better). An illustrative example of "ACC-CIFAR-10": Average accuracy metric on the 10-Split-CIFAR100 dataset with corresponding tuning parameters of 10 width.}
\label{fig_14}
\end{figure*}

We empirically investigate the proposed strategy via comprehensive experiments and sensitivity analyses, and demonstrate its generalization ability in various backbones, including (i) ViT and (ii) CLIP. We show that our method substantially outperforms prior state-of-the-art techniques across these backbones, which is observed in Figure \ref{fig_14}. Our contributions are summarized as follows:

\textbullet ~ Parameter efficient gradient projection is the first unified work to study the anti-forgetting mechanism of parameter-efficient tuning. Our approach can be applied in various parameter-efficient tuning paradigms, where the forgetting is significantly reduced on most benchmarks.

\textbullet ~Based on the hypothesis that old tasks should have the same results after model updating, our approach obtains the orthogonal condition of gradient and finds the projection matrix towards the direction that has less impact on the old feature space, with less extra memory space and training time.

\textbullet ~ Our approach achieves state-of-the-art results in terms of forgetting and average accuracy metrics, under the settings of task, domain, class, online class, and cross-modality incremental learning on diverse datasets.

\section{Related Work}
\subsection{Continual Learning}
Continual learning refers to training deep neural networks (DNNs) on time-variant data \cite{Continual1, Continual2, Continual3}. In this context, data is organized into a sequence of tasks denoted as \{$\mathcal{D} = \mathcal{D}_1, ..., \mathcal{D}_T$\}, where the $t$-th task $\mathcal{D}_t=\{(x_i^t, y_i^t)_{i=1}^{n_t}\}$ comprises input sample $x_i^t$ $\in$ $\mathcal{X}_t$ and its corresponding label $y_i^t$ $\in$ $\mathcal{Y}_t$. Upon arrival of a new task $\mathcal{D}_t$, a model $f$ is trained with no access to data from previous tasks.

In detail, it could be divided into \textit{i.e.}, task-, domain-, class- and online-incremental learning, abbreviated as TIL, DIL, CIL, and OIL respectively \cite{Three, OIL}. Providing a task identifier (abbreviated as task ID) in the inference stage is deemed as task incremental learning, otherwise class incremental learning. Each task maintains the same classes, while the domains of tasks are different from each other, which is recognized as domain incremental learning. Online class incremental learning refers to each training sample only appearing once, in other words, each task is trained with only a single epoch. In this paper, in order to verify the generalization of our method, the above incremental scenarios will be involved sequentially. 

\subsection{PET-Based Continual Learning}
PET-based continual learning originated from a simple yet effective Prompt-based CIL model: Learning to Prompt (L2P) \cite{L2P}. In it, a prompt pool existed, and an instance-wise query mechanism was enabled to pick up suitable prompts. Due to the repeated usage of prompts, newly learned knowledge would cover old ones, leading to forgetting. To address this problem, the S-Prompts method was proposed, utilizing task-specific prompts that were only trainable in the corresponding task and frozen in other tasks \cite{Sprompts}. However, assigning an independent prompt to each task introduces a memory burden as the number of tasks increases, and task identifier inference remains an unsolved challenge. To reduce forgetting while introducing fewer new prompts in each task, based on Prefix-tuning, DualPrompt divided the prompts into two parts: expert prompts (task-specific) and general prompts (task-shared), inserting them into different layers for learning distinct features \cite{DualPrompt}. 

One unified parameter-efficient continual method considered four tuning paradigms (Prompt-tuning, Prefix-tuning, Adapter, and LoRA) by a simple and direct Exponential Moving Average (EMA) method to resist forgetting \cite{LAE}. However, forgetting was not explicitly modeled in this framework, and the mechanism against forgetting has not been revealed yet.
 
In contrast to the above literature, our work is the first to propose a unified framework to solve the forgetting problem in PET-based continual learning with mathematical demonstration. Its theoretical deduction is in Sec.3 and its effectiveness is verified in Sec.4.

\begin{figure*}[t]
\centering
\includegraphics[width=5in]{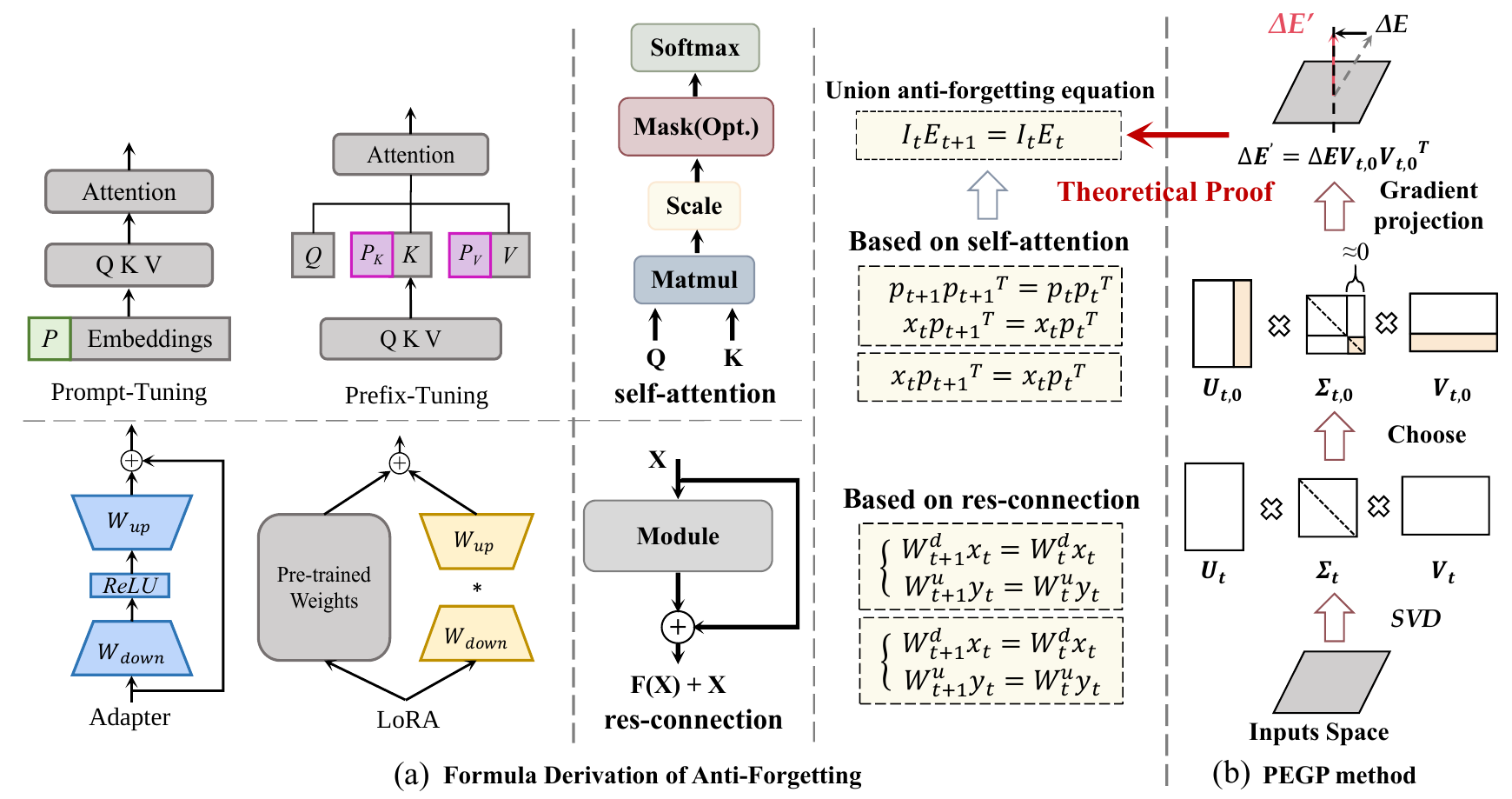}
\caption{Illustration of our motivations and methods. (a) Through the investigation of four PETs, we discover a unified anti-forgetting formula from two distinct mechanisms. (b) Implementation of the PEGP process, including feature space sampling, singular value decomposition, gradient projection matrix obtaining, and gradient projection.}
\label{fig_2}
\end{figure*}

\subsection{Background of Gradient Projection Method}

Gradient limitation, originating from sophisticated mathematical theory, restricts the gradient direction and offers a significant explanation for the stability-plasticity dilemma \cite{LOS, GEM, Decouple}. 

Recent research has demonstrated that learning can avoid forgetting if gradients are updated in the direction orthogonal to the subspace spanned by old features \cite{Subspaces}. The Gradient Projection Method (GPM), for instance, updates weights in the orthogonal direction to the subspace formed by previously learned inputs, thereby ensuring new learning processes do not disrupt prior tasks \cite{GPM}. Trust Region Gradient Projection (TRGP) extends this approach by constraining updates within a trust region and employing layer-wise scaling matrices along with orthogonal gradient projection to accommodate new tasks \cite{TRGP}. Simple Linear Connector (Connector) integrates two models using a weighted sum function, allowing one model to update normally while the other updates by using orthogonal gradient projection \cite{Connector}.

However, one limitation of gradient projection methods is only applicable to convolutional neural networks and lack a theoretical foundation in parameter-efficient tuning, which is based on vision transformers. In this paper, we will prove that gradient projection methods can aid in resisting forgetting for parameter-efficient tuning, and the combination of these two methods shows advanced properties in continual learning through experiments.

\subsection{Differences From the Preliminary Version}
This study is a journal extension of the conference paper with the following differences and improvements \cite{PGP}: 

(1) The motivations and formulations are different from our prior work \cite{PGP}. Our previous work primarily provides specific deductions and solutions to alleviate forgetting in Prompt/Prefix-tuning. In contrast, PEGP proposes a unified framework that can simultaneously solve the problem in total four parameter-efficient tuning paradigms (Adapter/LoRA/Prompt/Prefix-tuning). Additionally, PEGP migrates from single modality model, ViT, to cross-modality model, CLIP.

(2) The prerequisites are looser. In our prior work, we extra introduced key-query mechanism to provide task-identifier, which can be an aid in updating proper prompts with projected gradient. In this work, we discover that the gradient projection method can also work well without task identifier.

(3) We extensively evaluate PEGP across diverse continual learning scenes and provide comprehensive ablations isolating each key component. Notice that, we first construct a cross-modality continual learning benchmark called BITM, which is the task of image-text matching. Experiments across all scenarios verify both the theoretical soundness and practical effectiveness of our method.

\section{Method}
\subsection{A Unified Parameter-Efficient Continual Method} 
In our gradient projection framework, parameter-efficient tunings could be categorized as two sub-types: 1) Prompt/Prefix-tuning based on \textbf{self-attention} as prepended vectors interacting with inputs in self-attention calculation and 2) Adapter/LoRA based on \textbf{res-connection}, due to adding bypass in the net module. Although starting from distinct insertion, we surprisingly find that the forward of these modules seems similar and the anti-forgetting character can be described by a union equation. Here, we take class incremental learning as an example, and suppose we have trained on the $t$-th task and met the next $t+1$-th task. The old inputs are denoted as $x_t$ and the new/old trainable parameters as $E_{t+1}$, $E_t$ respectively. Our motivation and implementation method are shown in Figure \ref{fig_2}.

To better preserve old knowledge, we propose that in the ideal state of anti-forgetting, the update of the network would satisfy the following proposition:

\noindent{\textbf{Proposition 1. }}{\textit{Starting from the old inputs from previous tasks $x_t$ have the same outputs after learning a new task, we have:
\begin{align}
f_\theta(E_{t+1},x_t)=f_\theta(E_t,x_t),
\end{align}
}}
where $f_\theta$ refers to the backbone parameters. Unrelated with the specific tuning paradigm, forgetting is avoided if we can have the following equation (Detailed proof is in the following subsections):
\begin{align}\label{union}
x_tE_{t+1} = x_tE_t.
\end{align}
Expand $E_{t+1}$ as $E_{t+1}=E_t+\Delta E$ and remove the identical items in each side of the union equation. Finally, we can obtain:
\begin{align}\label{union}
x_t\Delta E = 0.
\end{align}
Eq.(\ref{union}) implies that if the gradient is updated in the orthogonal direction to the subspace spanned by old features, the forgetting would be significantly reduced.

\textbf{Discussion:} Compared with previous works, we firstly propose a union anti-forgetting method based on parameter efficient tunings. Please kindly refer to C in the supplementary materials for the detailed algorithm process. Our method is independent of the specific tuning mechanism and owns strong generalization ability without any assumptions.

\subsection{Self-Attention Based Gradient Projection Method}
\begin{figure}[h]
\centering
\includegraphics[width=2.7in]{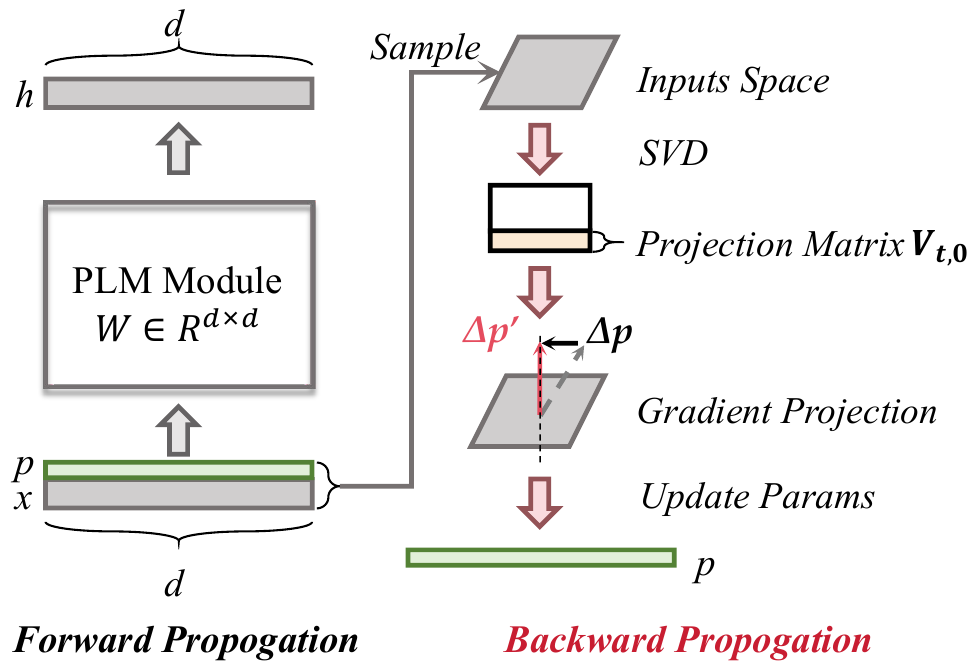}
\caption{Flowchart for Prompt-based gradient projection.}
\label{fig_3}
\end{figure}

\begin{figure}[h]
\centering
\includegraphics[width=2.7in]{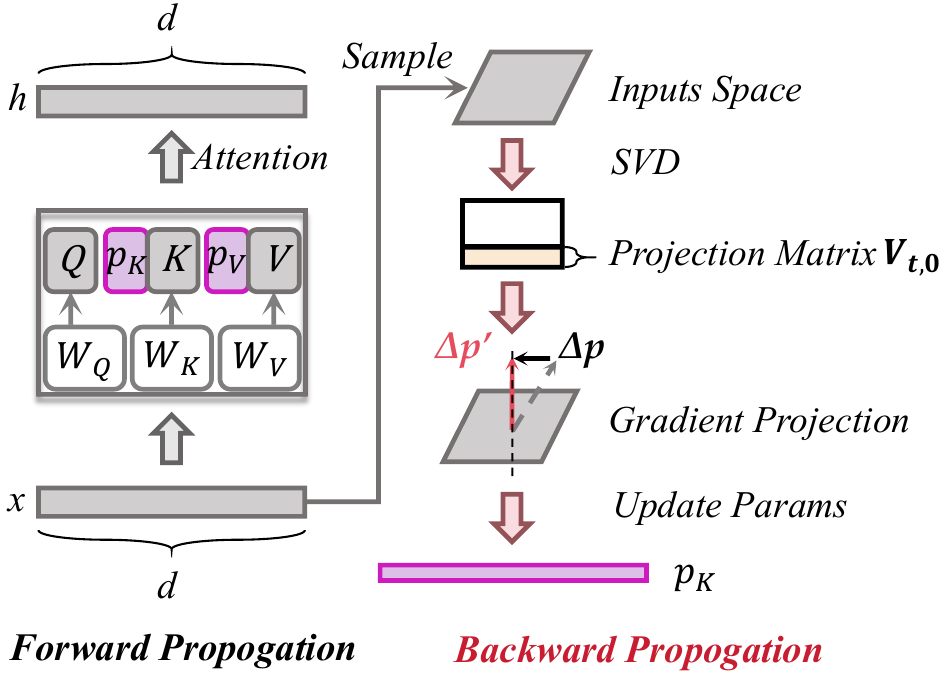}
\caption{Flowchart for Prefix-based gradient projection.}
\label{fig_4}
\end{figure}

Prompt-tuning prepends prompt vector $P\in{R^{l\times{d}}}$ ($l$ denotes the sequence length and $d$ denotes the embedding dimension) with the input embeddings $x_t$ in the first layer \cite{Prompt}. In this framework, after training task $t+1$, we concatenate the prompts $p_{t+1}$ and the embedding sequences $x_t$, {\it i.e.,} inputs from $t$-th task, along the embedding dimension: $Z_t^{t+1} = \begin{bmatrix} p_{t+1} \\ x_t \end{bmatrix}$. Thus the self-attention is conducted based on embedding $Z$, where query and key can be represented as $Q_t^{t+1}=W_qZ_t^{t+1}$ and $K_t^{t+1}=W_kZ_t^{t+1}$ respectively, which is shown in Figure \ref{fig_3}.

Similar to Prompt-tuning, Prefix-tuning prependes the prompt vectors ${\mathit{p}}$ $\in$ $R^{L_{\mathit{p}} \times D}$ with key $K$ and value $V$ as $[p^{k}; K]$ and $[p^{v}; V]$ respectively \cite{Prefix}. Assuming that a set of prefixes have been trained at task $t+1$, we input samples from task $t$ and have $Q_{t}^{t+1} = W_{q}x_{t}$, $K_{t}^{t+1} = \begin{bmatrix} p_{t+1}^k \\ W_{k}x_{t} \end{bmatrix}$, which is shown in Figure \ref{fig_4}. 

In order to realize Proposition 1, we start from the implementation of the self-attention matrix \cite{ViT}, which is calculated as:
\begin{align}
\label{Attention}
A_t^{t+1} = softmax(\frac{Q_t^{t+1}{K_t^{t+1}}^T}{\sqrt{(\frac{d}{h})}}).
\end{align}
Due to the normalized denominator, we mainly focus on the numerator part $Q_t^{t+1}{K_t^{t+1}}^T$. In the next, we will discuss two cases separately.

\textbf{Prompt-tuning:} The numerator in Eq.(\ref{Attention}) can be further expanded as $W_qZ_t^{t+1}{Z_t^{t+1}}^TW_k^T$. Because we freeze the weights of visual encoder, $W_q$ and $W_k$ are unchanged during training. The parameters influenced by training are $Z_t^{t+1} \cdot {Z_t^{t+1}}^T$.

To realize Proposition 1, {\it i.e.,} the condition of anti-forgetting, we need to achieve $Z_t^{t+1} \cdot {Z_t^{t+1}}^T=Z_t^{t} \cdot {Z_t^{t}}^T$ and the requirements are:
\begin{equation}\label{condition}
\left\{
\begin{aligned}
p_{t+1}p_{t+1}^T = p_{t}p_{t}^T,\\
x_tp_{t+1}^T = x_tp_{t}^T,\\
p_{t+1}x_t^T = p_{t}x_t^T.\\
\end{aligned}
\right.
\end{equation}
Please kindly refer to A.1 in the supplementary materials for the whole demonstration process of Eq.(\ref{condition}).

\textbf{Prefix-tuning:} By further expanding the numerator part $Q_t^{t+1}{K_t^{t+1}}^T$, we have the equation that satisfy Proposition 1 as:
\begin{equation}
\label{prefix_condi}
x_{t}p_{t+1}^T = x_{t}p_{t}^T,
\end{equation}
Please kindly refer to A.2 in the supplementary materials for the whole demonstration process of Eq.(\ref{prefix_condi}).

In the following content, we take the Prompt-tuning as an example. In Eq.(\ref{condition}), $p_{t+1}$ is divided into $p_t$ and $\Delta p$, where $\Delta p$ is the gradient of prompts when training task $t+1$\footnote{Here we omit the factor of learning rating since this simplification wouldn't influence our conclusion.}. Thus we extend $p_{t+1}p_{t+1}^T$ as:
\begin{align}
p_{t+1}p_{t+1}^T &= (p_t+\Delta p)(p_t+\Delta p)^T \\
&= p_tp_t^T+p_t\Delta p^T+\Delta pp_t^T+\Delta p\Delta p^T.
\end{align}
Here we ignore the high-order infinitesimal term of $\Delta p\Delta p^T$. By observation, if $p_t\Delta p^T=0$, the first term in Eq.(\ref{condition}) {\it i.e.,} $p_{t+1}p_{t+1}^T = p_{t}p_{t}^T$ can be realized.

Similarly, the second term in Eq.(\ref{condition}) can be expanded as:
\begin{align}
\label{important}
x_tp_{t+1}^T = x_t(p_{t}^T + \Delta p^T) = x_tp_{t}^T + x_t\Delta p^T = x_tp_{t}^T. 
\end{align}
Eliminating $x_tp_{t}^T$ on both sides, we have $x_t\Delta p^T = 0$. Note that this condition also satisfies the third term in Eq.(\ref{condition}) because $x_tp_{t+1}^T$ is the transpose of $p_{t+1}x_t^T$.

Therefore, we reach the key conclusion: anti-forgetting ability can be obtained by restricting the gradient of prompts to satisfy the following equations:
\begin{equation}\label{gradient_condi}
\left\{
\begin{aligned}
x_t\Delta p^T=0,\\
p_t\Delta p^T=0.\\
\end{aligned}
\right.
\end{equation}
To solve Eq.(\ref{gradient_condi}), we take the first term as an example, we decompose $x_t$ with SVD: $ x_t = U_t \Sigma_t V_t^{T}$. Here, $U_t$ and $V_t$ contain singular vectors corresponding to singular values in $\Sigma_t$, and matrix $\Sigma_t$ can be further divided as:
\begin{equation}
\Sigma_t=\begin{bmatrix}\Sigma_{t,1} & O \\ O & \Sigma_{t,0} \end{bmatrix},
\end{equation}
where $\Sigma_{t,1}$ denotes the non-zero singular values of $\Sigma_t$ and $\Sigma_{t,0}$ represents the near-zero singular values of $\Sigma_t$ \cite{SVD}. Corresponding to the value of element in $\Sigma_t$ whether equals to zero, $V_t$ can be divided into two parts along the column dimension: $V_t=[V_{t,1}, V_{t,0}]$. Thus, we have:
\begin{equation}
    x_{t}[V_{t,1}, V_{t,0}] = U_{t}\begin{bmatrix}\Sigma_{t,1} & O \\ O & \Sigma_{t,0} \end{bmatrix}.
\end{equation}
Finally, we have the equation:
\begin{equation}
    x_{t}V_{t,0} = U_{t}\begin{bmatrix}O \\ \Sigma_{t,0} \end{bmatrix} \approx O.
\end{equation}
Let $\Delta p=\Delta pV_{t,0}V_{t,0}^T$, we can obtain:
\begin{align}\label{projection}
x_{t}\Delta p^T = x_{t}{(\Delta pV_{t,0}V_{t,0}^T)}^T = x_{t}V_{t,0}V_{t,0}^T\Delta p^T = O.
\end{align}
By taking $V_{t,0}$ as the gradient projection matrix, we successfully realize the first term in Eq.(\ref{gradient_condi}).

\begin{figure}[h]
\centering
\includegraphics[width=2.7in]{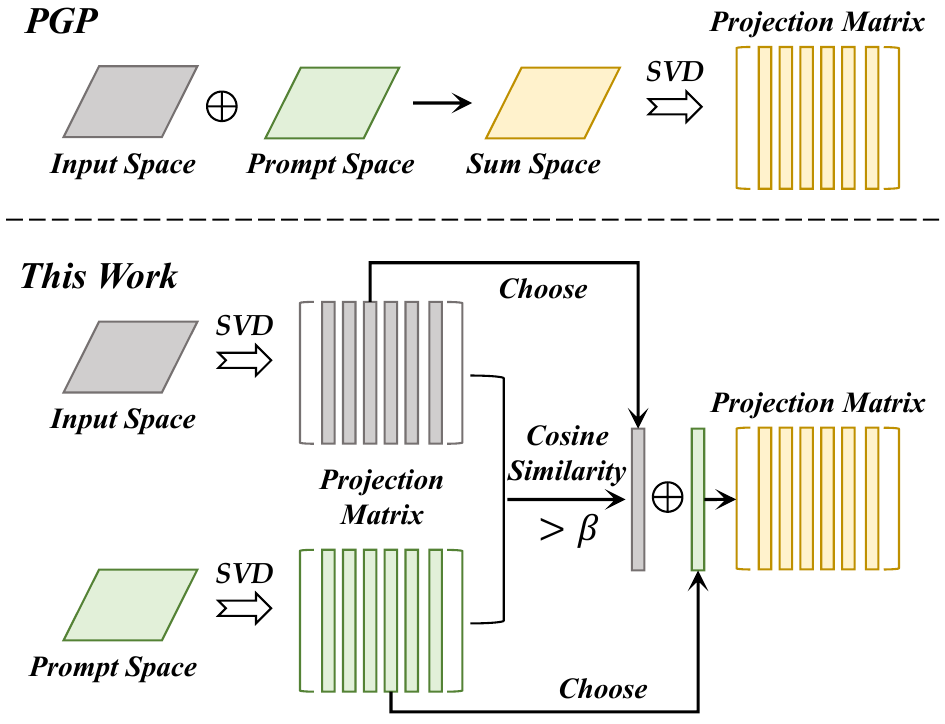}
\caption{Comparison of gradient projection matrix obtaining method between PGP and this work.}
\label{fig_16}
\end{figure}

In our previous work, we sum the prompt space and inputs space directly to create the sum space \cite{PGP}, on which we calculate the gradient projection matrix. However, the summation strategy would fail if the column vectors from each space own totally different directions, influencing the result of gradient projection.

In this paper, we calculate the cosine similarity between the column vectors of $V_{t,0}^i$ and $V_{t,0}^p$ in the same location, where $V_{t,0}^i$ denotes the projection matrix of inputs and $V_{t,0}^p$ refers to the projection matrix of prompt. Then we compare the similarity score with a hyperparameter threshold $\beta$. If the score is larger \textit{i. e.}, the two column vectors have similar directions, we sum the two column vectors and put the result into the final projection matrix $V_{t,0}$. Otherwise, we continue to compare the residual column vectors. The whole process is shown in Figure \ref{fig_16}.

\subsection{Res-Connection Based Gradient Projection Method}

\begin{figure}[h]
\centering
\includegraphics[width=3in]{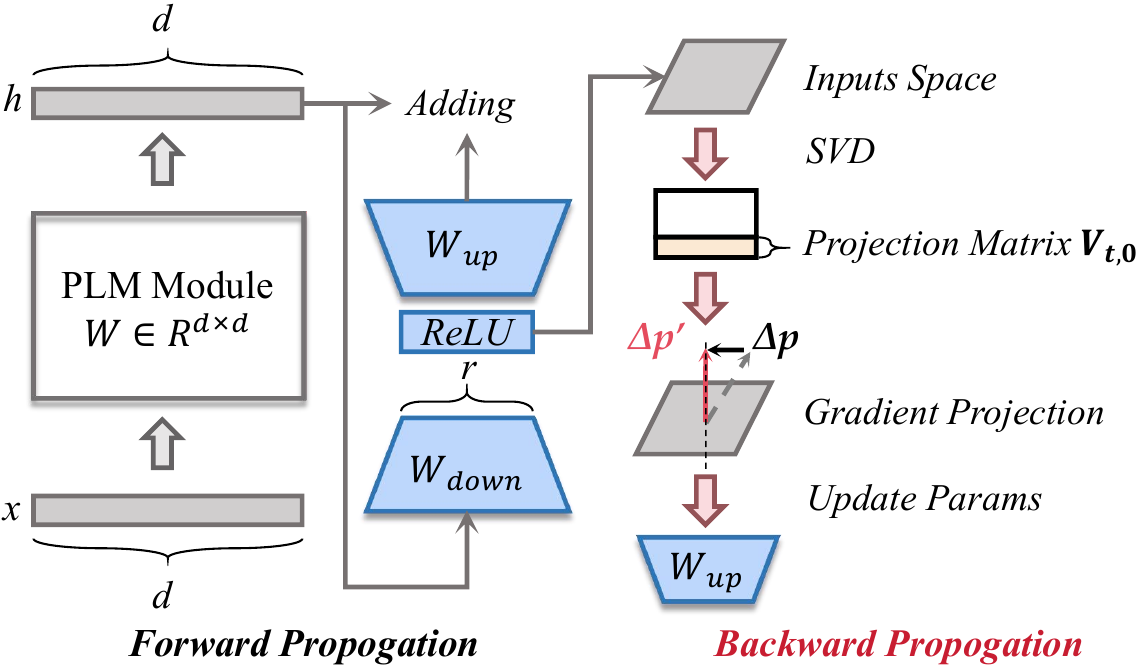}
\caption{Flowchart for Adapter-based gradient projection.}
\label{fig_5}
\end{figure}

\begin{figure}[h]
\centering
\includegraphics[width=3in]{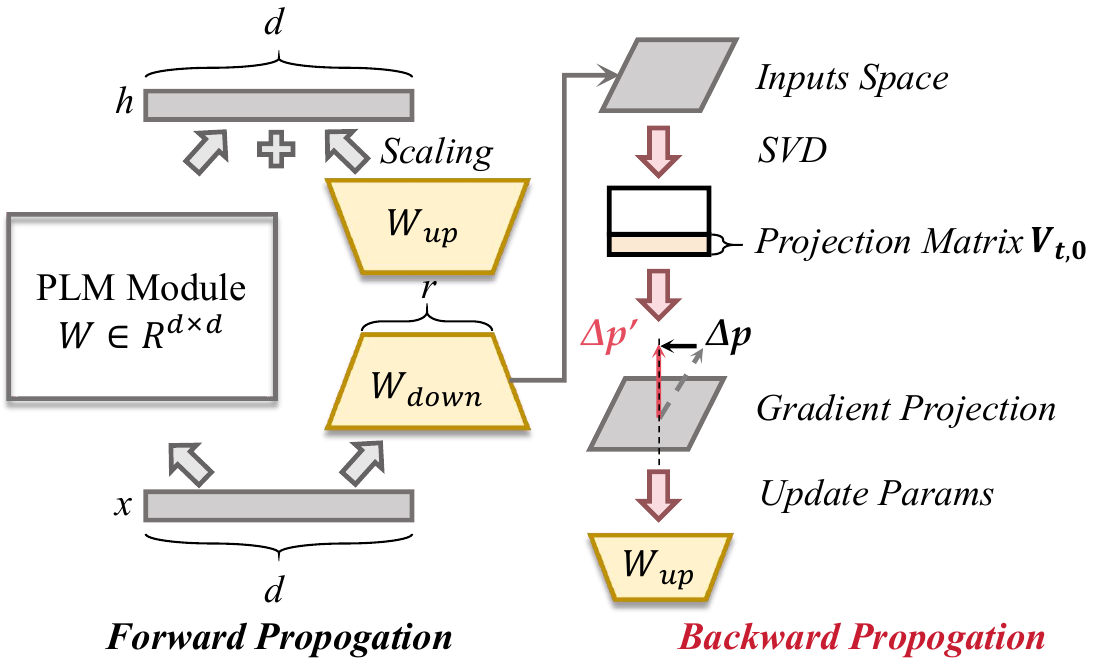}
\caption{Flowchart for LoRA-based gradient projection.}
\label{fig_6}
\end{figure}

Adapter includes a down-sampling layer with wights $W^d\in{R^{d\times{r}}}$ to reduce dimension from a higher dimension $d$ to a lower dimension $r$, followed by a nonlinear activation function $f(\cdot)$, and a up-sampling layer with weights $W^u\in{R^{r\times{d}}}$ to restore dimension from $r$ to $d$ \cite{Adapter}, which is shown in Figure \ref{fig_5}. In each Adapter module, inputs $x$ and outputs $h$ can be described as:
\begin{align}
h = f_{\theta}(x) + f_{act}(W^{u}W^{d}x).
\end{align}

LoRA prepends tiny trainable low-rank matrices $W^d\in{R^{d\times{r}}}$ and $W^u\in{R^{r\times{d}}}$ into backbone parameters as $W + \Delta W = W + W^uW^d$ \cite{LoRA}, which is shown in Figure \ref{fig_6}. In LoRA module, inputs $x$ and outputs $h$ can be described as:
\begin{align}
h = f_{\theta}(x) + s\cdot W^{u}W^{d}x.
\end{align}

From Proposition 1, if we eliminate identical terms on both sides of these two equations respectively, we have:
\begin{align}
\label{Adapter/LoRA condi}
\left\{
\begin{aligned}
f_{act}(W_t^{u} W_t^{d} x_t)=f_{act} (W_{t+1}^{u} W_{t+1}^{d} x_t),\\
s{\cdot}W_t^{u} W_t^{d} x_t=s{\cdot}W_{t+1}^{u} W_{t+1}^{d} x_t.\\
\end{aligned}
\right.
\end{align}
In addition to the scalar $s$ and activation function $f_{act}$, we can find that the mathematical form of Adapter is highly similar to LoRA. Thus, we can both obtain the following equation to resist the forgetting in both paradigms:
\begin{align}
\label{lora1}
W_t^{u}W_t^{d}x_t = W_{t+1}^{u}W_{t+1}^{d}x_t.
\end{align}
Define the update of $W^{d}$, $W^{u}$ as $\Delta W^{d}$ and $\Delta W^{u}$. We can present the $W^{d}_{t+1}$ and $W^{u}_{t+1}$ by $W^{d}_{t}$ and $W^{u}_{t}$ as:
\begin{align}
\label{lora2}
\left\{
\begin{aligned}
W_{t+1}^{d} = W_t^{d} + \Delta W^{d},\\
W_{t+1}^{u} = W_t^{u} + \Delta W^{u}.\\
\end{aligned}
\right.
\end{align}
By substituting Eq.(\ref{lora2}) into Eq.(\ref{lora1}), we can obtain:
\begin{align}
W_t^{u}W_t^{d}x_t = (W_t^{d} + \Delta W^{d})(W_t^{u} + \Delta W^{u})x_t.
\end{align}
By rearranging the above equations, it can be derived from:
\begin{align}
\label{lora3}
\left\{
\begin{aligned}
W_{t+1}^{d}x_t = (W_t^{d} + \Delta W^{d})x_t,\\
W_{t+1}^{u}y_t = (W_t^{u} + \Delta W^{u})y_t,\\
\end{aligned}
\right.
\end{align}
where $y_t=W_t^{d}x_t=W_{t+1}^{d}x_t$. Removing the identical items in the both sides of Eq.(\ref{lora3}), we can yield:
\begin{align}
\label{lora4}
\left\{
\begin{aligned}
\Delta W^{d}x_t = 0,\\
\Delta W^{u}y_t = 0.\\
\end{aligned}
\right.
\end{align}
Therefore, Proposition 1 is shifted to Eq.(\ref{lora4}), where two equations have the same form.

To achieve Eq.(\ref{lora4}), taking $\Delta W^{d}x_t = 0$ as an example, we perform singular value decomposition and obtain $SVD(x_t)=U_t\Sigma_t{V_t}^T$. Further we have ${U_t}^Tx_t=\Sigma_t{V_t}^T$. By selecting only the eigenvectors in $U_t$ corresponding to zero or close to zero eigenvalues, we can obtain:
\begin{align}
{U_{t,0}}^Tx_t = \Sigma_{t,0}{V_{t,0}}^T = \begin{bmatrix}O \\ \Sigma_{t,0} \end{bmatrix} {V_{t,0}}^T \approx O.
\end{align}
Thus, if we choose ${U_{t,0}}{U_{t,0}}^T$ as the projection matrix, then we have: $\Delta W^{d} = \Delta W^{d}{U_{t,0}}{U_{t,0}}^T$. At this point, it holds that:
\begin{align}\label{adapter_projection}
\Delta W^{d}x_t = \Delta W^{d}{U_{t,0}}{U_{t,0}}^Tx_t = 0.
\end{align}
Similarly, we can also achieve $\Delta W^{u}y_t = 0$.


\section{Experiments}
In this section, we show the proposed PEGP framework could be flexibly applied to various continual learning tasks: (i) Class/Online class Incremental Learning; (ii) Domain Incremental Learning; (iii) Task Incremental Learning; (iv) Cross-modality Incremental Learning with distinct backbones (i) ViT; (ii) CLIP. The quantitative and qualitative results demonstrate the effectiveness of our approach in various circumstances. All experiments are held on NVIDIA RTX 4090 GPUs.

\subsection{Evaluation Benchmarks and Protocol}
In this section, we introduce the adopted benchmark datasets and corresponding evaluation standards.

\textbf{10-Split-CIFAR100 \cite{CIFAR100}} contains 20 major categories and 100 subcategories. For each subcategory, it owns 600 images (500 training images and 100 test images). We construct it by evenly splitting the 100 classes into 10 disjoint tasks, and each task has 10 classes. 

\textbf{10-Split-ImageNet-R \cite{IMR}} contains art, cartoons, deviantart, et al. 16 renditions of ImageNet classes. ImageNet-R has 200 ImageNet classes with a total of 30,000 images. We split the total 200 classes into 10 disjoint tasks, and each task has 20 classes.

\textbf{10-Split-ImageNet100 \cite{ImageNet}} selects 100 categories from ImageNet1K. The training set contains 600 annotated images for each class, and the validation set contains 100 annotated images for each class. We split the total 100 classes into 10 disjoint tasks, and each task has 10 classes.

\textbf{5-Split-DomainNet \cite{DomainNet}} is a dataset for domain adaption and domain incremental learning, which has 345 categories and roughly 600,000 images. Images from DomainNet are split into 5 domains. Here, we deem each domain as a task and each task has 345 classes.

We follow the most popular protocol for evaluation, where average accuracy (Simplified as accuracy or Avg. Acc), forgetting, and new task accuracy (Simplified as New Acc) (please kindly refer to B in the supplementary materials for more details).

\subsection{Implementation Details}
\textbf{For ViT Backbone:} Consistent with previous works, we use ViT-B/16 pre-trained on ImageNet-21K as our image encoder \cite{ViT}, which is kept frozen during training. We adopt the LAE model, which is a continual learning method containing Adapter/LoRA/Prompt/Prefix-tuning paradigms, and remove the original online/offline dual model design and EMA model fusion mechanism as our baseline. Furthermore, we add the proposed parameter efficient gradient projection method at the training stage to suppress forgetting.

\textbf{For CLIP Backbone:} We utilize the model from our previous work \cite{PGP}, and the backbone is ViT-B-16 pre-trained by OpenAI \cite{CLIP}. On the vision side, we only set a single trainable image Prompt/Linear Adapter shared by all tasks. As for the text side, we set trainable text prompt for each class, which is only trained at the corresponding task according to CoOP \cite{CoOP}. Besides that, we add the gradient projection method at the training stage for efficient parameters and improve our gradient projection method by following \cite{Connector}, which linearly merges two models with/without gradient projection method \cite{Linear}. 

\begin{table*}[thb]
\small
\caption{Class incremental learning (\textit{i.e.}, task identifier is unknown at test phase) results of Avg. ACC and Forgetting on 10-Split-CIFAR100 and 10-Split-ImageNet-R with ViT backbone.}
\label{Table 2}
\centering
\scalebox{0.9}{
\begin{tabular}{c|c|c||cc||cc}
\toprule 
 \multirow{2}{*}{\textbf{Method}} & \multirow{2}{*}{\textbf{Avenue}} & \multirow{2}{*}{\textbf{Paradigm}} & \multicolumn{2}{c||}{\textbf{10-Split-CIFAR100}} & \multicolumn{2}{c}{\textbf{10-Split-ImageNet-R}} \\
& & & Avg. Acc ($\uparrow$) & Forgetting ($\downarrow$) & Avg. Acc ($\uparrow$) & Forgetting ($\downarrow$) \\
\midrule
 &  & Adapter-5 & 79.96 & 12.60 & 62.23 & 18.58 \\
 &  & Adapter-10 & 79.17 & 13.81 & 61.98 & 19.71 \\
 &  & LoRA-5 & 78.26 & 12.66 & 61.19 & 17.13 \\
 &  & LoRA-10 & 79.45 & 12.60 & 61.49 & 18.27 \\
 &  & Prompt-10 & 78.92 & 9.88 & 58.98 & 11.77 \\
 &  & Prompt-20 & 80.13 & 9.32 & 59.37 & 11.71 \\
 &  & Prefix-10 & 77.04 & 11.32 & 58.95 & 13.13 \\
\multirow{-8}{*}{Baseline} & \multirow{-8}{*}{-} & Prefix-20 & 77.68 & 11.80 & 60.08 & 12.98 \\
\midrule
L2P\cite{L2P} & CVPR\color{blue}{'22} & Prompt & 83.12 & 7.66 & 68.38 & 6.93 \\
\midrule
DualPrompt\cite{DualPrompt} & ECCV\color{blue}{'22} & Prefix & 84.59 & 5.60 & 68.57 & 6.29 \\
\midrule
 &  & Adapter-5 & 84.75 & 7.43 & 70.72 & 9.62 \\
 &  & Adapter-10 & 85.10 & 6.69 & 70.82 & 9.89 \\
 &  & LoRA-5 & 84.80 & 6.64 & 69.65 & 10.13 \\
 &  & LoRA-10 & 85.04 & 6.91 & 69.44 & 10.92 \\
 &  & Prompt-10 & 84.23 & 5.80 & 68.58 & 6.12 \\
 &  & Prompt-20 & 84.06 & 6.32 & 68.81 & 6.38 \\
 &  & Prefix-10 & 85.03 & 5.97 & 70.86 & 6.28 \\
\multirow{-8}{*}{LAE\cite{LAE}} & \multirow{-8}{*}{ICCV\color{blue}{'23}} & Prefix-20 & 84.83 & 6.37 & 70.51 & 6.59 \\ 
\midrule
\rowcolor{LightCyan} &  & Adapter-5 & 85.16 & 6.09 & 71.59 & 5.62 \\
\rowcolor{LightCyan} &  & Adapter-10 & 85.43 & 6.38 & 71.70 & 5.85 \\
\rowcolor{LightCyan} &  & LoRA-5 & 85.02 & 6.54 & 70.30 & 6.35 \\
\rowcolor{LightCyan} &  & LoRA-10 & 85.14 & 6.23 & 70.54 & 6.04 \\
\rowcolor{LightCyan} &  & Prompt-10 & 84.69 & 5.47 & 69.30 & 5.94 \\
\rowcolor{LightCyan} &  & Prompt-20 & 84.46 & 6.11 & 69.31 & 6.10 \\
\rowcolor{LightCyan} &  & Prefix-10 & 85.35 & 5.75 & 71.04 & 6.09 \\
\rowcolor{LightCyan} \multirow{-8}{*}{\textbf{PEGP}} & \multirow{-8}{*}{\textbf{Ours}} & Prefix-20 & 85.30 & 6.11 & 70.79 & 6.27 \\
\bottomrule
\end{tabular}}
\end{table*}

\subsection{Class/Online Class/Task Incremental Learning}
\subsubsection{Experiments on ViT Backbone}
\textbf{Training Settings:} We train the 10-Split-CIFAR100 for 5 epochs and 10-Split-ImageNet-R for 50 epochs with 24 images (resized as 224*224*3) in each batch. Adapter/LoRA width is set at 5/10, and Prefix/Prompt length is set at 10/20. The initial learning rate is 0.0028125 for 10-Split-CIFAR100 and 0.00046875 for 10-Split-ImageNet-R, and the decay rate is 0 with Adam optimizer \cite{CE, Softmax, Adam}.

\begin{table}[h]
\small
\caption{Online class incremental learning results of Avg. ACC and Forgetting on 10-Split-CIFAR100 dataset with ViT backbone.}
\label{Table 3}
\centering
\scalebox{0.85}{
\begin{tabular}{c|c|c||cc}
\toprule 
 \multirow{2}{*}{\textbf{Method}} & \multirow{2}{*}{\textbf{Avenue}} & \multirow{2}{*}{\textbf{Paradigm}} & \multicolumn{2}{c}{\textbf{10-Split-CIFAR100}} \\
& & & Avg. Acc ($\uparrow$) & Forgetting ($\downarrow$) \\
\midrule
Baseline & - & Adapter-5 & 75.19 & 19.61 \\
Baseline & - & LoRA-5    & 76.58 & 16.00 \\
\midrule
LAE\cite{LAE} & ICCV\color{blue}{'23} & Adapter-5 & 81.11 & 11.44 \\
LAE\cite{LAE} & ICCV\color{blue}{'23} & LoRA-5    & 78.30 & 12.17 \\
\midrule
\rowcolor{LightCyan} \textbf{Ours} & - & \textbf{Adapter-5} & \textbf{83.77} & \textbf{6.82} \\
\rowcolor{LightCyan} \textbf{Ours} & - & \textbf{LoRA-5}   & \textbf{83.21} & \textbf{7.47} \\
\bottomrule
\end{tabular}}
\end{table}

\textbf{Experimental Results:} We compare the performance of our PEGP method with other SOTA methods in Table \ref{Table 2} under the class incremental setting. We observe that PEGP can greatly improve the average accuracy and reduce forgetting compared with the baseline on the four tuning paradigms, demonstrating its effectiveness and generalization ability. It is also noteworthy that with the aid of gradient projection, PEGP outperforms other SOTA methods on the two datasets by +0.33@Avg. ACC on the 10-Split-CIFAR100 benchmark and +0.88@Avg. ACC on the 10-Split-ImageNet-R benchmark.

\begin{figure}[htb]
\centering
\includegraphics[width=3.5in]{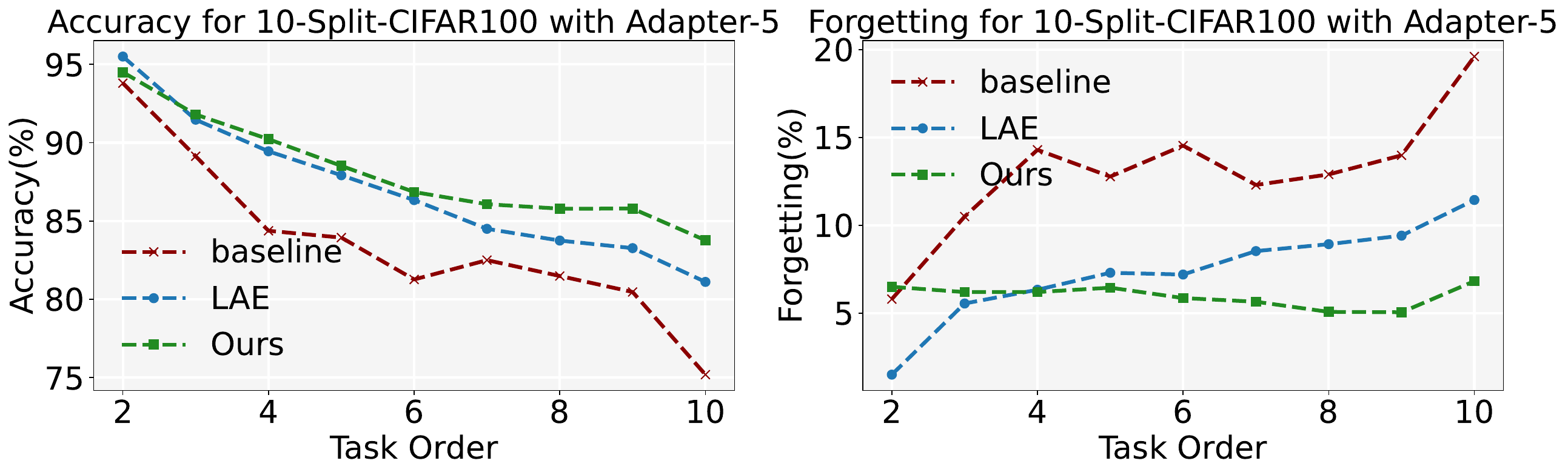}
\caption{Task-by-task performance changing curves in terms of accuracy and forgetting under online class incremental learning on 10-Split-CIFAR100 with Adapter-5.}
\label{fig_8}
\end{figure}

\begin{figure}[htb]
\centering
\includegraphics[width=3.5in]{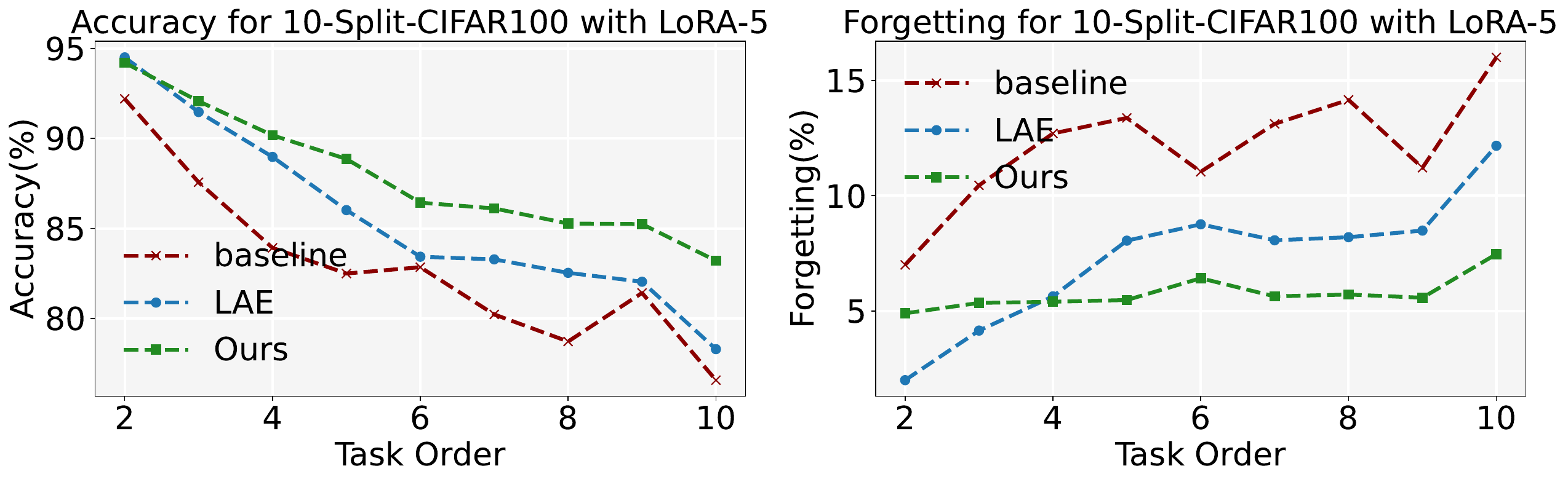}
\caption{Task-by-task performance changing curves in terms of accuracy and forgetting under online class incremental learning on 10-Split-CIFAR100 with LoRA-5.}
\label{fig_9}
\end{figure}

The same experimental phenomena can also be explored under online class incremental learning in Table \ref{Table 3}. Although the training process only allows one epoch for each task \cite{Privacy}, our method still performs stronger anti-forgetting ability compared with baseline and others. Figure \ref{fig_8} and Figure \ref{fig_9} show the curves of accuracy and forgetting with the task number increasing on 10-Split-CIFAR100. We observe that on all tasks, the accuracy of our method is always higher than baseline, and forgetting is always lower than baseline with two different tuning paradigms.

\begin{figure*}[hbt]
\centering
\includegraphics[width=5in]{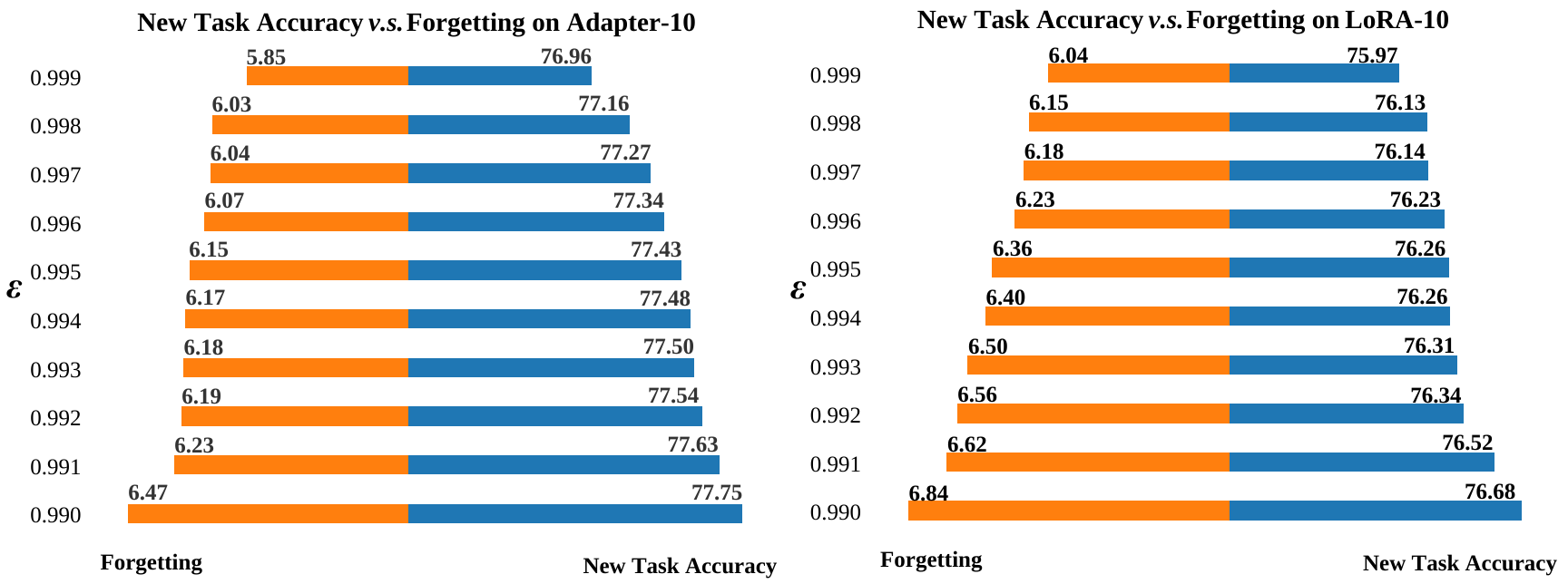}
\caption{Ablation study with threshold $\epsilon$ with Adapter-10/LoRA-10 paradigm on 10-Split-ImageNet-R.}
\label{fig_11} 
\end{figure*}

\textbf{Ablation Study:} Based on the Figure \ref{fig_11}, we can draw similar conclusions in Adapter/LoRA tuning case: With the change of hyper-parameter $\epsilon$, leading to the prompt projection matrix $V_{t,0}$ with distinct numbers of column vectors. If the $\epsilon$ becomes smaller, more column vectors would be added into the $V_{t,0}$, causing lower new task accuracy (worse plasticity ) but less forgetting (better stability). On the contrary, bigger $\epsilon$ refers to the fewer column vectors in the $V_{t,0}$, resulting in more forgetting (worse stability) but higher new task accuracy (better plasticity).

Although here we only show the results with Adapter/LoRA tuning paradigms, similar phenomena are also observed in our previous work with Prompt/Prefix tuning paradigms, and detailed mathematical mechanisms behind them are also included \cite{PGP}. Thus, We can recognize that the essence of the gradient projection method is a kind of trade-off strategy between plasticity and stability. However, different from other dilemmas \cite{Dilemma}, it has an optimal solution, which is projecting gradient in the direction orthogonal to the subspace spanned by the old inputs, which can not only own the best ability of anti-forgetting but also have minimal damage to plasticity.

\subsubsection{Experiments on CLIP Backbone}
\textbf{Training Settings:} We train the 10-Split-CIFAR100/10- Split-ImageNet100 for 5 epochs with 32 images (resized as 224*224*3) in each batch. Linear Adapter width/Prompt length is set at 10. The initial learning rate is 0.001 for the first task and 0.01 for sequential tasks, while the decay rate is 0.1 with the Adam optimizer.

\begin{table*}[bth]
\small
\caption{Class incremental learning results on 10-Split-CIFAR100 and 10-Split-ImageNet100 dataset with CLIP backbone.}
\label{Table 4}
\centering
\scalebox{0.9}{
\begin{tabular}{c|c|c||cc||cc}
\toprule 
 \multirow{2}{*}{\textbf{Method}} & \multirow{2}{*}{\textbf{Avenue}} & \multirow{2}{*}{\textbf{Paradigm}} & \multicolumn{2}{c||}{\textbf{10-Split-CIFAR100}} & \multicolumn{2}{c}{\textbf{10-Split-ImageNet100}} \\
& & & Avg. Acc ($\uparrow$) & Forgetting ($\downarrow$) & Avg. Acc ($\uparrow$) & Forgetting ($\downarrow$) \\
\midrule
Continual-CLIP\cite{continual-clip} & ArXiv\color{blue}{'22} & Prompt & \multicolumn{1}{c|}{66.70} & - & \multicolumn{1}{c|}{75.40} & - \\
\midrule
CoOP(Baseline)\cite{CoOP} & IJCV\color{blue}{'22} & Prompt & \multicolumn{1}{c|}{73.76} & 5.60 & \multicolumn{1}{c|}{83.66} & 2.47 \\
\midrule
CoOP(Baseline)\cite{CoOP} & IJCV\color{blue}{'22} & Adapter & \multicolumn{1}{c|}{63.95} & 6.28 & \multicolumn{1}{c|}{78.24} & 5.04 \\
\midrule
CLIP-EWC\cite{EWC} & PNAS\color{blue}{'17} & Prompt & \multicolumn{1}{c|}{72.29} & 6.31 & \multicolumn{1}{c|}{81.92} & 3.58 \\
\midrule
CLIP-LWF\cite{LWF} & TPAMI\color{blue}{'16} & Prompt & \multicolumn{1}{c|}{75.44} & 7.42 & \multicolumn{1}{c|}{84.10} & 2.38 \\
\midrule
CLIP-PGP\cite{PGP} & ICLR\color{blue}{'24} & Prompt & \multicolumn{1}{c|}{79.47} & 4.23 & \multicolumn{1}{c|}{84.14} & 2.11 \\
\midrule
AttriCLIP\cite{AttriCLIP} & CVPR\color{blue}{'23} & Prompt & \multicolumn{1}{c|}{81.40} & - & \multicolumn{1}{c|}{83.30} & - \\
\midrule
\rowcolor{LightCyan} \textbf{CLIP-PEGP} & \textbf{Ours} & \textbf{Prompt} & \multicolumn{1}{c|}{\textbf{82.36 (+8.60)}} & \textbf{2.82 (-2.78)} & \multicolumn{1}{c|}{\textbf{85.54 (+1.88)}} & \textbf{1.80 (-0.67)} \\
\midrule
\rowcolor{LightCyan} \textbf{CLIP-PEGP} & \textbf{Ours} & \textbf{Adapter} & \multicolumn{1}{c|}{70.37 (+6.42)} & 5.72 (-0.56) & \multicolumn{1}{c|}{79.16 (+0.92)} & 2.60 (-2.44) \\
\bottomrule
\end{tabular}}
\end{table*}

\begin{table*}[thb]
\small
\caption{Task incremental learning results on 10-Split-CIFAR100 and 10-Split-ImageNet100 dataset with CLIP backbone.}
\label{Table 5}
\centering
\scalebox{0.9}{
\begin{tabular}{c|c|c||cc||cc}
\toprule 
 \multirow{2}{*}{\textbf{Method}} & \multirow{2}{*}{\textbf{Avenue}} & \multirow{2}{*}{\textbf{Paradigm}} & \multicolumn{2}{c||}{\textbf{10-Split-CIFAR100}} & \multicolumn{2}{c}{\textbf{10-Split-ImageNet100}} \\
& & & Avg. Acc ($\uparrow$) & Forgetting ($\downarrow$) & Avg. Acc ($\uparrow$) & Forgetting ($\downarrow$) \\
\midrule
CoOP(Baseline)\cite{CoOP} & IJCV\color{blue}{'22} & Prompt & \multicolumn{1}{c|}{92.69} & 2.34 & \multicolumn{1}{c|}{89.08} & 1.98 \\
\midrule
CLIP-EWC\cite{EWC} & PNAS\color{blue}{'17} & Prompt & \multicolumn{1}{c|}{\textbf{94.42}} & 1.22 & \multicolumn{1}{c|}{89.08} & 1.60 \\
\midrule
CLIP-LWF\cite{LWF} & TPAMI\color{blue}{'16} & Prompt & \multicolumn{1}{c|}{93.96} & 1.38 & \multicolumn{1}{c|}{\textbf{89.36}} & \textbf{1.56} \\
\midrule
CLIP-PGP\cite{PGP} & ICLR\color{blue}{'24} & Prompt & \multicolumn{1}{c|}{93.00} & 1.58 & \multicolumn{1}{c|}{88.75} & 1.73 \\
\midrule
\rowcolor{LightCyan} \textbf{CLIP-PEGP} & \textbf{Ours} & \textbf{Prompt} & \multicolumn{1}{c|}{93.99} & \textbf{0.80} & \multicolumn{1}{c|}{89.08} & \textbf{1.56} \\
\bottomrule
\end{tabular}}
\end{table*}
\textbf{Experimental Results:} As we compare our method with other SOTA methods based on CLIP backbone, we see that the PEGP can also greatly bring decent improvements both on average accuracy and forgetting, as shown in Table \ref{Table 4}. Specifically, on 10-Split-CIFAR100, the method sees an improvement of +0.96 in average accuracy compared with the SOTA method. Similarly, on 10-Split-ImageNet100, the method sees an improvement of surprisingly +1.40 in average accuracy compared with the SOTA method. This validates that the introduction of gradient projection method with parameter-efficient tuning improves model performance without requiring any additional learnable parameters.

\begin{figure}[htb]
\centering
\includegraphics[width=3.5in]{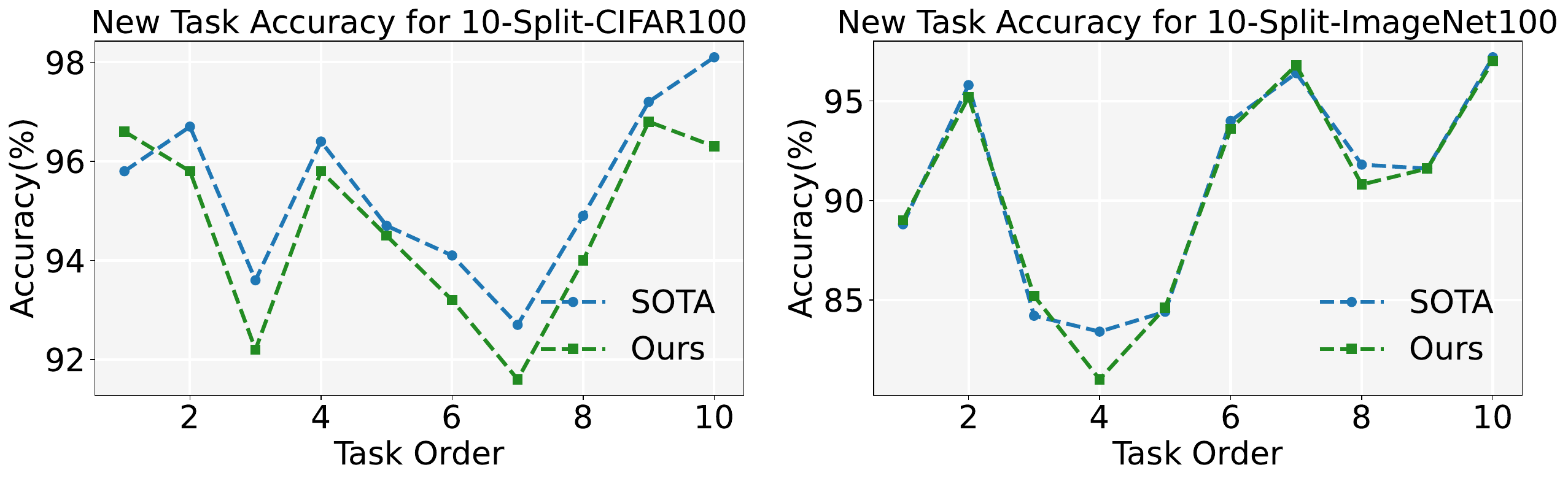}
\caption{New task accuracy changing curves with CLIP.}
\label{fig_10}
\end{figure}

\begin{figure*}[htb]
\centering
\includegraphics[width=5in]{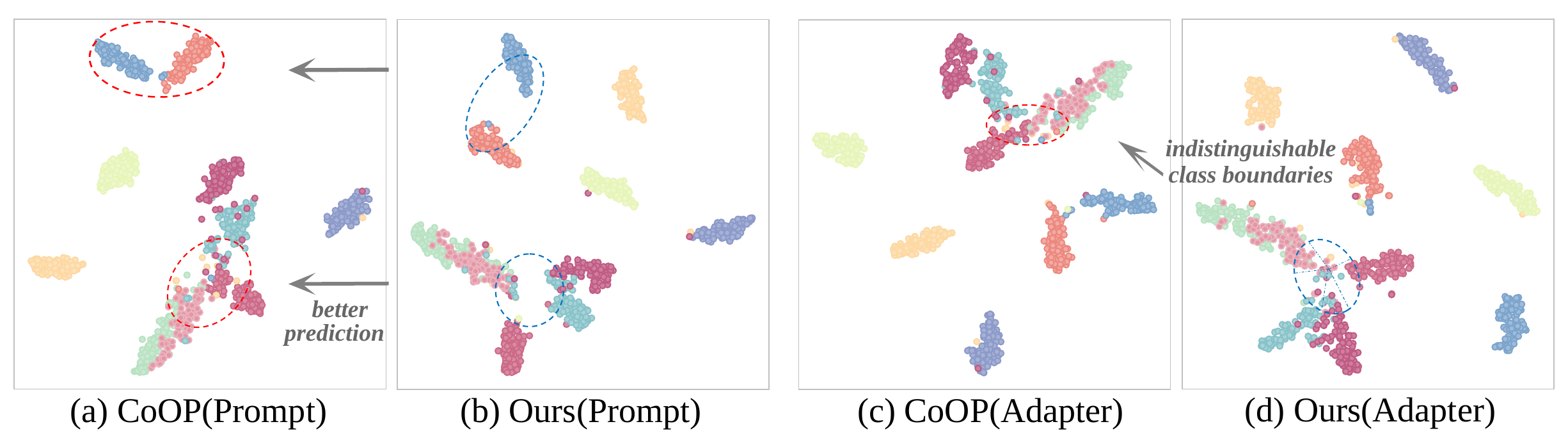}
\caption{t-SNE visualization of class incremental learning results on 10-Split-CIFAR100 based on CLIP backbone.}
\label{fig_12}
\end{figure*}

\begin{figure}[htb]
\centering
\includegraphics[width=3.5in]{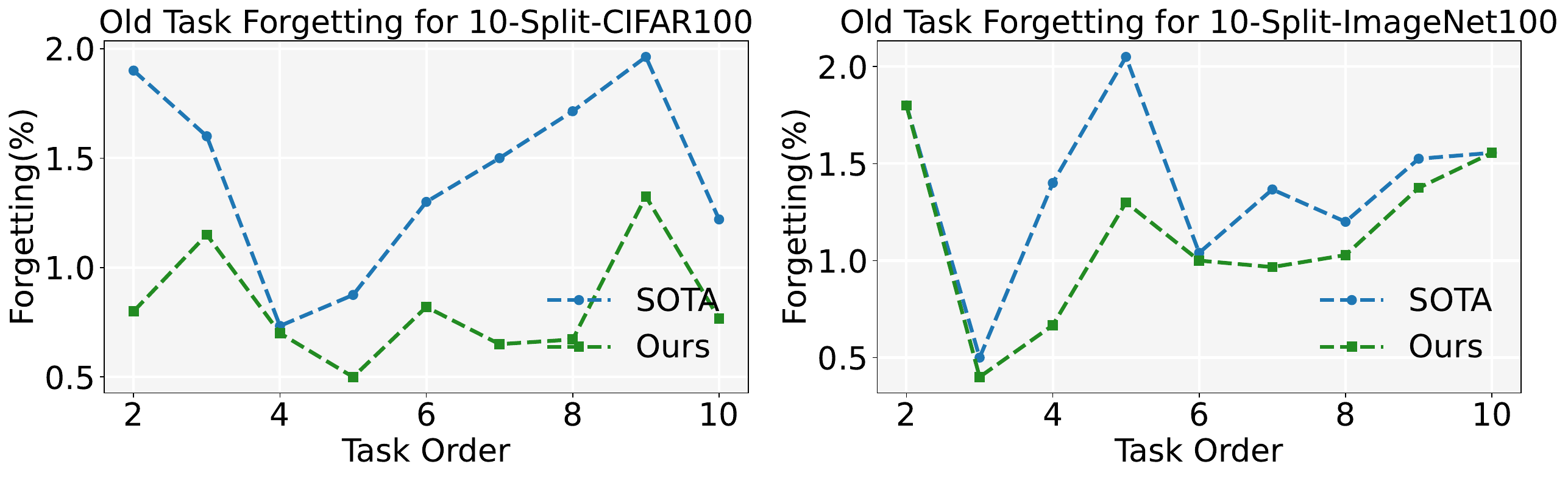}
\caption{Old task Forgetting changing curves with CLIP.}
\label{fig_15}
\end{figure}

While for task incremental setting, experiments show that our method owns the lowest forgetting but slightly lower accuracy than state-of-the-art on both two datasets in Table \ref{Table 5}. Reasons we speculate is that plasticity and stability are trade-off in continual learning. If we focus more on maintaining stability, plasticity is inevitably affected, resulting in a decrease in overall average accuracy. To prove our above hypothesis, we present the new class accuracy changing and old task forgetting curves for each of the 10 tasks, as shown in Figure \ref{fig_10} and Figure \ref{fig_15}. It can be seen that the overall new task accuracy curve of our method is always lower than the SOTA method, while old task forgetting is also.

\begin{table*}[thb]
\small
\caption{Domain incremental learning results of Avg. ACC and Forgetting on DomainNet dataset with ViT backbone.}
\label{Table 6}
\centering
\scalebox{0.9}{
\begin{tabular}{c|c||cc||cc||cc}
\toprule 
 \multirow{2}{*}{\textbf{Method}} & \multirow{2}{*}{\textbf{Width}} & \multicolumn{2}{c||}{\textbf{Baseline}} & \multicolumn{2}{c||}{\textbf{LAE\cite{LAE}}} & \multicolumn{2}{c}{\textbf{Ours}} \\
& & Avg. Acc ($\uparrow$) & Forgetting ($\downarrow$) & Avg. Acc ($\uparrow$) & Forgetting ($\downarrow$) & Avg. Acc ($\uparrow$) & Forgetting ($\downarrow$) \\
\midrule
\multirow{2}{*}{\textbf{Adapter}} & 5 & 61.56 & 14.68 & 65.65 & 9.73 & {\cellcolor{LightCyan}} \textbf{66.30} & {\cellcolor{LightCyan}} \textbf{6.57} \\
 & 10 & 61.40 & 16.23 & 65.85 & 10.38 & {\cellcolor{LightCyan}} \textbf{67.05} & {\cellcolor{LightCyan}} \textbf{6.84} \\
 \midrule
\multirow{2}{*}{\textbf{LoRA}} & 5 & 61.59 & 14.92 & 66.13 & 9.68 & {\cellcolor{LightCyan}} \textbf{67.12} & {\cellcolor{LightCyan}} \textbf{5.88} \\
 & 10 & 61.80 & 15.51 & 66.09 & 10.77 & {\cellcolor{LightCyan}} \textbf{67.17} & {\cellcolor{LightCyan}} \textbf{6.46} \\
 \midrule
\multirow{2}{*}{\textbf{Prefix}} & 10 & 61.56 & 14.68 & 65.87 & 8.36 & {\cellcolor{LightCyan}} \textbf{66.03} & {\cellcolor{LightCyan}} \textbf{7.64} \\
 & 20 & 61.46 & 13.99 & 65.80 & 8.95 & {\cellcolor{LightCyan}} \textbf{66.25} & {\cellcolor{LightCyan}} \textbf{8.13} \\
 \midrule
\multirow{2}{*}{\textbf{Prompt}} & 10 & 61.34 & 11.99 & 64.74 & 7.28 & {\cellcolor{LightCyan}} \textbf{64.89} & {\cellcolor{LightCyan}} \textbf{5.98} \\
 & 20 & 61.41 & 12.28 & 64.82 & 7.43 & {\cellcolor{LightCyan}} \textbf{65.23} & {\cellcolor{LightCyan}} \textbf{6.38} \\
\bottomrule
\end{tabular}}
\end{table*}

\textbf{Visualization:} To better visualize the improvement of our method, we also plot the 2D projection of representations, obtained from the visual output of CLIP. As shown in Figure 12, by using t-SNE, we compare the representations obtained from our approach and baseline. The representations are from task 1 and processed by model trained in task 9. By observing the scatters, we have: (i) the feature space embedded by baseline seems to lack discrimination, where we can observe the overlapping phenomenon between different classes; (ii) the boundaries between some classes of the feature space embedded by baseline looks indistinguishable. However, the above two problems have been well addressed in our method, suggesting better continual learning ability and old knowledge preservation.

\begin{table}[h]
\caption{Zero-shot performance with CLIP backbone. The column of CLIP refers to the zero-shot performance of original CLIP.}
\label{Table 8}
\centering
\scalebox{1.0}{
\begin{tabular}{c|c|c|c}
\toprule 
 {\textbf{Method}} & {\textbf{Baseline}} & {\textbf{PEGP}} & {\textbf{CLIP}} \\
\midrule
CIFAR100$\rightarrow$ImageNet100 & 70.56 & \textbf{72.84} & \color{red}75.42 \\
ImageNet100$\rightarrow$CIFAR100 & 52.85 & \textbf{59.91} & \color{red}66.72 \\
\bottomrule
\end{tabular}}
\end{table}

\textbf{Zero Shot Performance:} Catastrophic forgetting can severely impact the zero-shot generalization ability of large-scale pre-trained models, leading to the zero-shot collapse phenomenon. Based on the proposed PEGP, we design experiments to validate its ability to mitigate zero-shot collapse in pre-trained models. Utilizing CLIP, we first continually train on the 10-Split-CIFAR100 dataset, followed by testing zero-shot inference ability on the ImageNet100 dataset. Reversely, we continually train on the 10-Split-ImageNet100 dataset and test zero-shot inference ability on the CIFAR100 dataset. The experimental results, as presented in Table \ref{Table 8}, demonstrate that PEGP significantly enhances the zero-shot generalization ability of the pre-trained model after continuous fine-tuning in downstream tasks compared to the baseline.

\subsection{Domain Incremental Learning}

Domain incremental learning aims to verify the continual domain adaption ability of methods. In this setting, images in each task come from distinct domains and the number and kinds of classes in each task should remain the same. This scenario is therefore satisfactory to validate the anti-forgetting effectiveness of our PEGP method with changing domains.

\textbf{Training Settings:} We only adopt the ViT backbone in this setting and train the 5-Split-DomainNet for 50 epochs with 24 images (resized as 224*224*3) in each batch. Adapter/LoRA width is set at 5/10, and Prefix/Prompt length is set at 10/20. The initial learning rate is 0.00046875, and the decay rate is 0 with the Adam optimizer.

\textbf{Experimental Results:} Table \ref{Table 6} shows the comparison results of our PEGP method with baseline and LAE. We observe from Table \ref{Table 6} that PEGP again sets a new state-of-the-art in this setting too. Compared to 5-Split-DomainNet where our method improves SOTA by +1.20@Avg. ACC and -3.54@Forgetting on the Adapter-10 paradigm at the most with the aid of gradient projection, demonstrating its excellent anti-forgetting and generalization ability for the four PET methods.

\begin{table*}[t]
\small
\caption{Cross-modality incremental learning results with CLIP backbone on BITM dataset.}
\label{Table 7}
\centering
\scalebox{0.9}{
\begin{tabular}{c|c|c||cc||cc}
\toprule 
 \multirow{2}{*}{\textbf{Method}} & \multirow{2}{*}{\textbf{Avenue}} & \multirow{2}{*}{\textbf{Paradigm}} & \multicolumn{2}{c||}{\textbf{10-Split-BITM}} & \multicolumn{2}{c}{\textbf{5-Split-BITM}} \\
& & & Avg. Acc ($\uparrow$) & Forgetting ($\downarrow$) & Avg. Acc ($\uparrow$) & Forgetting ($\downarrow$) \\
\midrule
CoOP(Baseline)\cite{CoOP} & IJCV\color{blue}{'22} & Prompt & \multicolumn{1}{c|}{42.01} & 33.04 & \multicolumn{1}{c|}{52.73} & 29.32 \\
\midrule
CLIP-EWC\cite{EWC} & PNAS\color{blue}{'17} & Prompt & \multicolumn{1}{c|}{50.20} & 32.62 & \multicolumn{1}{c|}{37.03} & 25.82 \\
\midrule
CLIP-LWF\cite{LWF} & TPAMI\color{blue}{'16} & Prompt & \multicolumn{1}{c|}{55.81} & 23.48 & \multicolumn{1}{c|}{18.31} & 49.07 \\
\midrule
CLIP-PGP\cite{PGP} & ICLR\color{blue}{'24} & Prompt & \multicolumn{1}{c|}{54.26} & 23.01 & \multicolumn{1}{c|}{59.74} & 24.35 \\
\midrule
\rowcolor{LightCyan} \textbf{CLIP-PEGP} & \textbf{Ours} & \textbf{Prompt} & \multicolumn{1}{c|}{\textbf{55.70}} & \textbf{22.97} & \multicolumn{1}{c|}{\textbf{62.66}} & \textbf{23.38} \\
\bottomrule
\end{tabular}}
\end{table*}

\subsection{Cross-modality Incremental Learning}
\begin{figure}[h]
\centering
\includegraphics[width=2.5in]{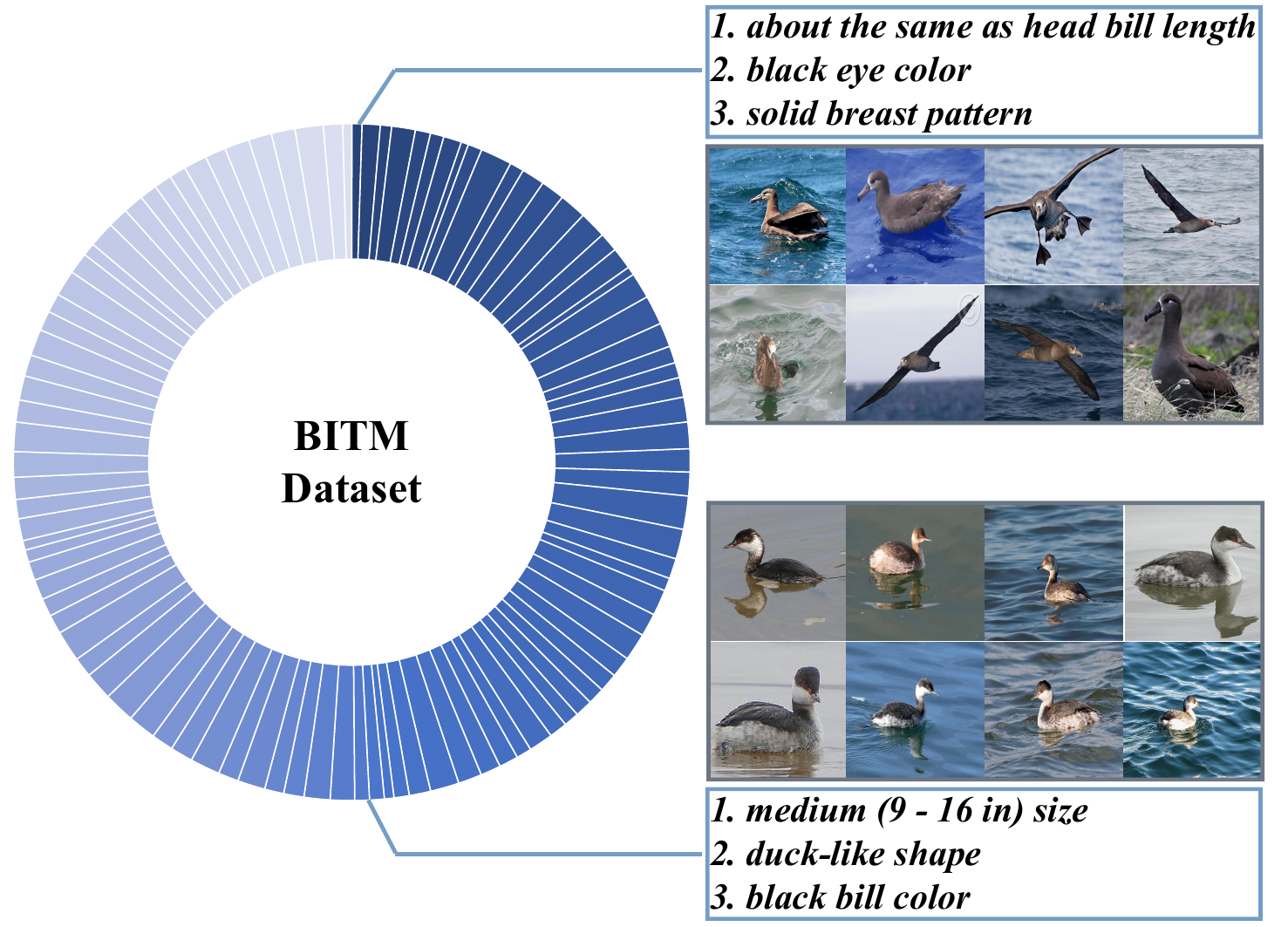}
\caption{Visualization of BITM: a novel text-image matching dataset. In the figure, each sector represents a class, and the area of the sector represents the number of images in the corresponding class. The larger the sector, the more images in the corresponding class}
\label{fig_7}
\end{figure}

Cross-modality incremental learning intends to evaluate the effectiveness of the method in continually learning across modalities involving image and text. In this paper, we propose a novel image-text matching dataset called BITM based on the CUB200 dataset \cite{CUB200}. The specific construction method is as follows:

First of all, we select the first 100 classes from the CUB200 dataset. Based on attribute annotations, we extract attributes with certainties of 4 (definitely) and 3 (probably) of each class. These attributes are then sorted based on their frequency of occurrence in images, and the top three attributes with the highest frequency are selected as the text labels. Next, we select images that simultaneously possessed these text labels from the image set. These selected images, along with the text labels, form the text-image matching dataset. We name this new dataset as BITM (Bird Image Text Matching Dataset) and analyze the distribution of text labels and images in the dataset. The visualization of the dataset is shown in the Figure \ref{fig_7}. The statistics reveal that BITM contains 100 bird species with a total of 3714 images. The class with the highest number of images contains 59 images, while the class with the lowest number of images contains 13 images. On average, each class contains about 35 images. We divide the dataset into training and testing sets in the ratio of 8:2. Considering that BITM is a continual learning dataset, we further split the dataset into 10-Split-BITM and 5-Split-BITM, where each task contains 10 and 20 classes of images, respectively.

\textbf{Benchmarks:} 5/10-Split-BITM are two benchmarks for cross-modality incremental learning. Images from BITM are split into 5/10 disjoint tasks with 20/10 classes in each task.

\textbf{Training Settings:} In this setting, we adopt CLIP backbone, incorporating Prompt and Linear Adapter, into the image encoder as baselines. Additionally, in our method, we add PEGP. We train the 5/10-Split-BITM for 30 epochs with 32 images (resized as 224*224*3) in each batch. Linear Adapter width/Prompt length is set at 10. The initial learning rate is 0.001 for the first task and 0.01 for sequential tasks, while the decay rate is 0.1 with the Adam optimizer.

\begin{figure}[htb]
\centering
\includegraphics[width=3.5in]{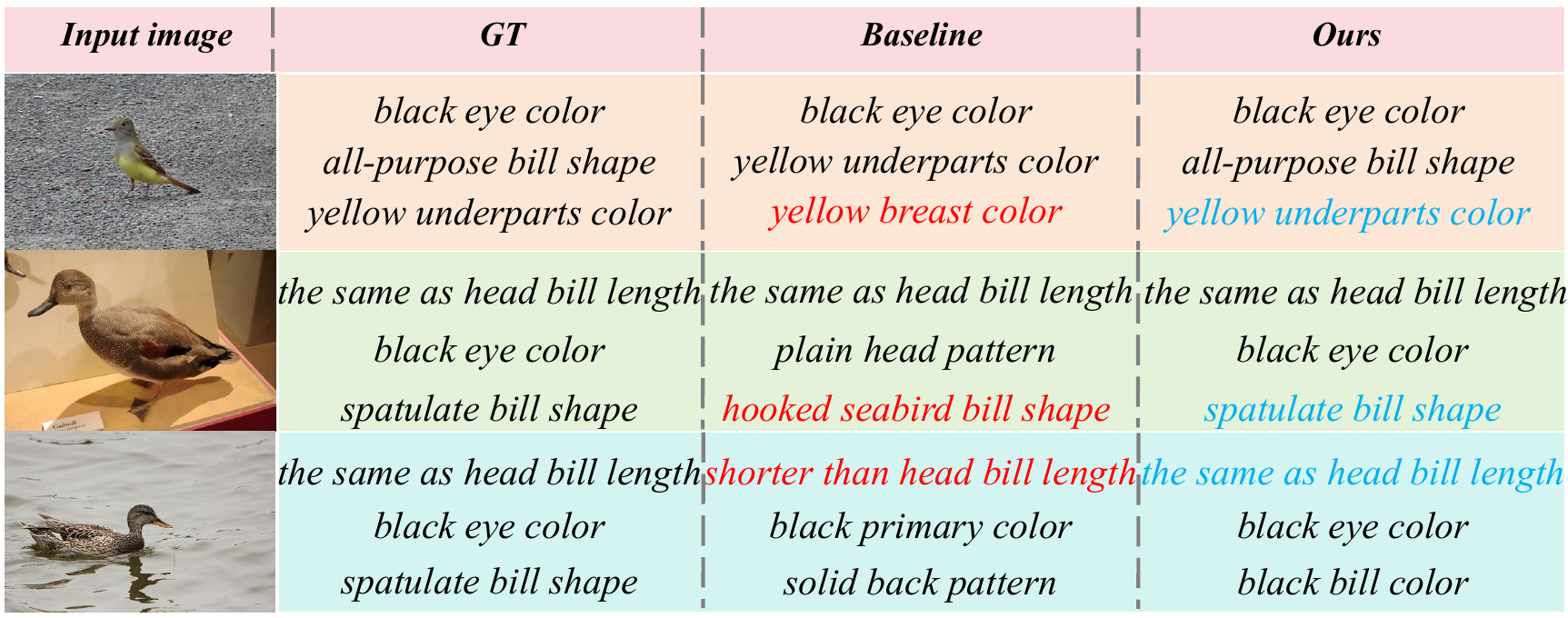}
\caption{Examples of cross-modality incremental learning results. The red-marked attributes represent the hallucinations in baseline, while the blue-marked attributes represent the predictions corresponding to our method.}
\label{fig_13}
\end{figure}

\textbf{Experimental Results:} We compare the performance of our PEGP method with baseline and other SOTA methods in Table \ref{Table 7}. We observe that PEGP can greatly improve the average accuracy and reduce forgetting compared with the baseline on the two benchmarks with distinct tuning paradigms, demonstrating its effectiveness and generalization ability. It is also noteworthy that with the aid of gradient projection, PEGP can not only reduce forgetting but also alleviate the hallucination appearing in the training process. In Figure \ref{fig_13}, we can clearly observe that hallucination is present during continually sequential fine-tuning. \cite{Hallucination} deem that the hallucination problem in large models is related to the forgetting problem in continual learning, and it can be grouped as external hallucination and internal hallucination. The above view is consistent with our experimental results. However, we are surprised to find that PEGP can spontaneously suppress the occurrence of hallucinations. We believe that this is inseparable from the anti-forgetting ability of our method. Therefore, we propose that mitigating illusions in large models can be examined through the perspective of reducing forgetting. This discovery may pave the way for addressing the issue of hallucination in large models.

\section{Conclusion}
In this work, we theoretically demonstrate that the orthogonal gradient projection can effectively resist forgetting and be applicable to all PET-based continual learning methods. On this basis, we provide a unified framework, Parameter Efficient Gradient Projection, to address the catastrophic forgetting for different tuning paradigms. Additionally, PEGP is proven to be the optimal solution for balancing the trade-off between plasticity and stability in PET-based continual learning methods by choosing the gradient projection matrix. Surprisingly, we also discover that the proposed method aids in overcoming the hallucination problem in the continual learning process. Future works would include learning the large language or multi-modality model to further verify the potential of the gradient projection method. Also, broader learning applications such as image generation or segmentation would be studied.

{\appendix[A. Proof Details]
\subsection*{A.1 Proof to Anti-forgetting Equations in Prompt-tuning}
In the Prompt-tuning paradigm, to realize Proposition 1, we have the following result from $Z_t^{t+1} \cdot {{Z_t^{t+1}}}^T$:
\begin{equation}
\label{appendix1}
\begin{split}
Z_t^{t+1} \cdot {Z_t^{t+1}}^T &= \begin{bmatrix}
					 	  p_{t+1} \\ x_t
					 	  \end{bmatrix} \begin{bmatrix}
						      p_{t+1}^T & x_t^T
						      \end{bmatrix} 
						      = \begin{bmatrix}
						      p_{t+1}p_{t+1}^T & p_{t+1}x_t^T \\
						      x_tp_{t+1}^T & x_tx_t^T 
						      \end{bmatrix}.
\end{split}
\end{equation}
By contrast, if we replace $p_{t+1}$ with $p_t$, we can obtain the old embedding $Z_t^t$ through calculation of concatenating prompt trained at task $t$ and embedding sequences $x_t$, and have:
\begin{equation}
\label{appendix2}
\begin{split}
Z_t^{t} \cdot {Z_t^{t}}^T &= \begin{bmatrix}
					 	  p_{t} \\ x_t
					 	  \end{bmatrix} \begin{bmatrix}
						      p_{t}^T & x_t^T
						      \end{bmatrix} 
						      = \begin{bmatrix}
						      p_{t}p_{t}^T & p_{t}x_t^T \\
						      x_tp_{t}^T & x_tx_t^T 
						      \end{bmatrix}.
\end{split}
\end{equation}
To achieve Proposition 1, {\it i.e.,} the condition of anti-forgetting, we need to make the equal of Eq.(\ref{appendix1}) and Eq.(\ref{appendix2}), and have:
\begin{equation}
\left\{
\begin{aligned}
p_{t+1}p_{t+1}^T = p_{t}p_{t}^T,\\
x_tp_{t+1}^T = x_tp_{t}^T,\\
p_{t+1}x_t^T = p_{t}x_t^T.\\
\end{aligned}
\right.
\end{equation}

\subsection*{A.2 Proof to Anti-forgetting Equations in Prefix-tuning}
With prefixes in key vector and trained at task $t+1$, we input samples from task $t$ and have:
\begin{equation}
Q_{t}^{t+1} = W_{q}x_{t},
\end{equation}
\begin{equation}
{K_{t}^{t+1}} = \begin{bmatrix}
p_{t+1} \\ W_{k}x_{t} \end{bmatrix},
\end{equation}
where, $W_{q}$ and $W_{k}$ are weights of $i$-th layer, frozen and unchanged. With Eq.(\ref{Attention}), our focus turns to the changing part:
\begin{align*}
Q_{t}^{t+1}{K_{t}^{t+1}}^T &= W_{q}x_{t}\begin{bmatrix} p_{t+1}^T & (W_{k}x_{t})^T \end{bmatrix} \\
&= \begin{bmatrix}W_{q}x_{t}p_{t+1}^T & W_{q}x_{t}x_{t}^TW_{k}^T\end{bmatrix}.
\end{align*}

Notice that $W_{q}x_{t}x_{t}^TW_{k}^T$ is stable, we only focus on the item $W_{q}x_{t}p_{t+1}^T$. Changing $p_{t+1}^T$ with $p_{t}^T$, we can obtain:
\begin{align*}
Q_{t}^{t}{K_{t}^{t}}^T &= W_{q}x_{t}\begin{bmatrix} p_{t}^T & (W_{k}x_{t})^T \end{bmatrix} \\
&= \begin{bmatrix}W_{q}x_{t}p_{t}^T & W_{q}x_{t}x_{t}^TW_{k}^T\end{bmatrix}.
\end{align*}

Considering that $W_{q}$ is the network parameter, which is frozen, our final goal can be simplified as the following equation to make the equal of $Q_{t}^{t+1}{K_{t}^{t+1}}^T$ and $Q_{t}^{t}{K_{t}^{t}}^T$.
\begin{equation}
x_{t}p_{t+1}^T = x_{t}p_{t}^T,
\end{equation}

\section*{B. Evaluation Metrics}
\textbf{Three metrics:} \textbf{Average Accuracy} (Avg. ACC), \textbf{Forgetting} (FOR), and \textbf{New Task Accuracy} (New. ACC) are used to evaluate the performance. We use average accuracy metric, for average test classification accuracy of all tasks. We adopt forgetting metric to indicate the loss of accuracy of past tasks after learning the new task. We employ new task accuracy metric, for average test classification accuracy of new tasks. 

\begin{equation}
\text{Average Accuracy} = \frac{1}{T}\sum_{i=1}^{T}A_{T,i},
\end{equation}
\begin{equation}
\text{Forgetting} = \frac{1}{T-1}\sum_{i=1}^{T-1}{A_{T,i} – \text{max}(A_{j,i})_{j \in [i,T-1]}},
\end{equation}
\begin{equation}
\text{New Task Accuracy} = \frac{1}{T}\sum_{i=1}^{T}A_{i,i},
\end{equation}
where $T$ is the number of tasks, $A_{T,i}$ is the accuracy of $i$-th task samples on the $T$-th model, $A_{j,i}$ is the accuracy of $i$-th task samples on the $j$-th model, and $A_{i,i}$ is the accuracy of $i$-th task samples on the $i$-th model.

\section*{C. Algorithm}
\begin{algorithm}
       \caption{\textbf{Gradient Projection Training Process}}
        \KwIn{ViT model $f_\theta$, classifier $f_c$, number of tasks $T$, training set ${\{\{x_i^t,y_i^t\}_{i=1}^{n_t}\}}_{t=1}^T$, sampling set ${\{\{x_{si}^t,y_{si}^t\}_{i=1}^{n_st}\}}_{t=1}^T$, efficient parameters $p$, number of training epochs $E$, projection matrix $V_{t,0}$, cross-entropy loss $\mathcal{L}_{\text{CE}}$, learning rate $\eta$}
        \textbf{initialize:} $f_c$, $p$ \\
       \For {$t$ = 1, ..., $T$}
       {
           \For{$e_1$ = 1, ..., $E$}
           {
               1: Draw a mini-batch $B$ = ${\{(x_i^t, y_i^t)\}}_{i=1}^{n_t}$ \\
               2: Obtain batch loss $\mathcal{L}_B$ by accumulating $ \mathcal{L}_{\text{CE}}(y_i^t,f_c(f_{\theta}(p, x_i^t)))$ \\
           }
           \# Parameter Efficient Gradient Projection \\
            \For {$t > 1$}
               {
                   3. Update $p$ by $p \leftarrow p - \eta \nabla_p\mathcal{L}_BV_{t,0}V_{t,0}^T$ with Eq.(\ref{projection}) or Eq.(\ref{adapter_projection}). \\
               }
	       \# Gradient Projection Matrix Update \\
           4. Initialize the sets of sampled features: $X_t=\{\}$. \\
           \For {${(x, y)}$ in ${\{(x_{si}^t, y_{si}^t)\}}_{i=1}^{n_{st}}$}
           {
               5. Sample set of features $X_t$ from $f_\theta(p, x)$. \\
               6. Update $V_{t,0}$ by $X_t$ according to \cite{PGP}. \\

           }
       }
\end{algorithm}

\vfill

\end{document}